\documentclass[pmlr]{jmlr}

\RequirePackage{graphicx}
 \usepackage{booktabs}
 \usepackage{multirow}
 \usepackage{rotating}
\usepackage{longtable}
\makeatletter
\def\set@curr@file#1{\def\@curr@file{#1}} 
\makeatother
\usepackage[load-configurations=version-1]{siunitx} 

\theorembodyfont{\upshape}
\theoremheaderfont{\scshape}
\theorempostheader{:}
\theoremsep{\newline}

\jmlrvolume{298}
\jmlryear{2024}
\jmlrworkshop{Machine Learning for Healthcare}

\title[The Impact of Image Resolution on Biomedical Multimodal Large Language Models]{The Impact of Image Resolution on Biomedical Multimodal Large Language Models}

\author{\Name{Liangyu Chen}
       \Email{liangyuc@stanford.edu}\\ 
       \addr Dept. of Computer Science\\
       Stanford University\\
       Stanford, 94305, CA 
       \AND
       \Name{James Burgess}
       \Email{jburgess@stanford.edu}\\ 
       \addr Institute for Computational and Mathematical Engineering (ICME)\\
       Stanford University\\
       Stanford, 94305, CA
       \AND
       \Name{Jeffrey Nirschl}
       \Email{jnirschl@stanford.edu}\\ 
       \addr Dept. of Pathology\\
       Stanford University\\
       Stanford, 94305, CA
       \AND
       \Name{Orr Zohar}
       \Email{orrzohar@stanford.edu}\\ 
       \addr Dept. of Electrical Engineering\\
       Stanford University\\
       Stanford, 94305, CA
       \AND
       \Name{Serena Yeung-Levy}
       \Email{syyeung@stanford.edu}\\ 
       \addr Dept. of Biomedical Data Science\\
       Stanford University\\
       Stanford, 94305, CA}

\begin{document}

\maketitle

\begin{abstract}
Imaging technologies are fundamental to biomedical research and modern medicine, requiring analysis of high-resolution images across various modalities. While multimodal large language models (MLLMs) show promise for biomedical image analysis, most are designed for low-resolution images from general-purpose datasets, risking critical information loss. We investigate how image resolution affects MLLM performance in biomedical applications and demonstrate that: (1) native-resolution training and inference significantly improve performance across multiple tasks, (2) misalignment between training and inference resolutions severely degrades performance, and (3) mixed-resolution training effectively mitigates misalignment and balances computational constraints with performance requirements. Based on these findings, we recommend prioritizing native-resolution inference and mixed-resolution datasets to optimize biomedical MLLMs for transformative impact in scientific research and clinical applications.
\end{abstract}

\begin{figure}[t]
  \centering
  \includegraphics[width=\linewidth]{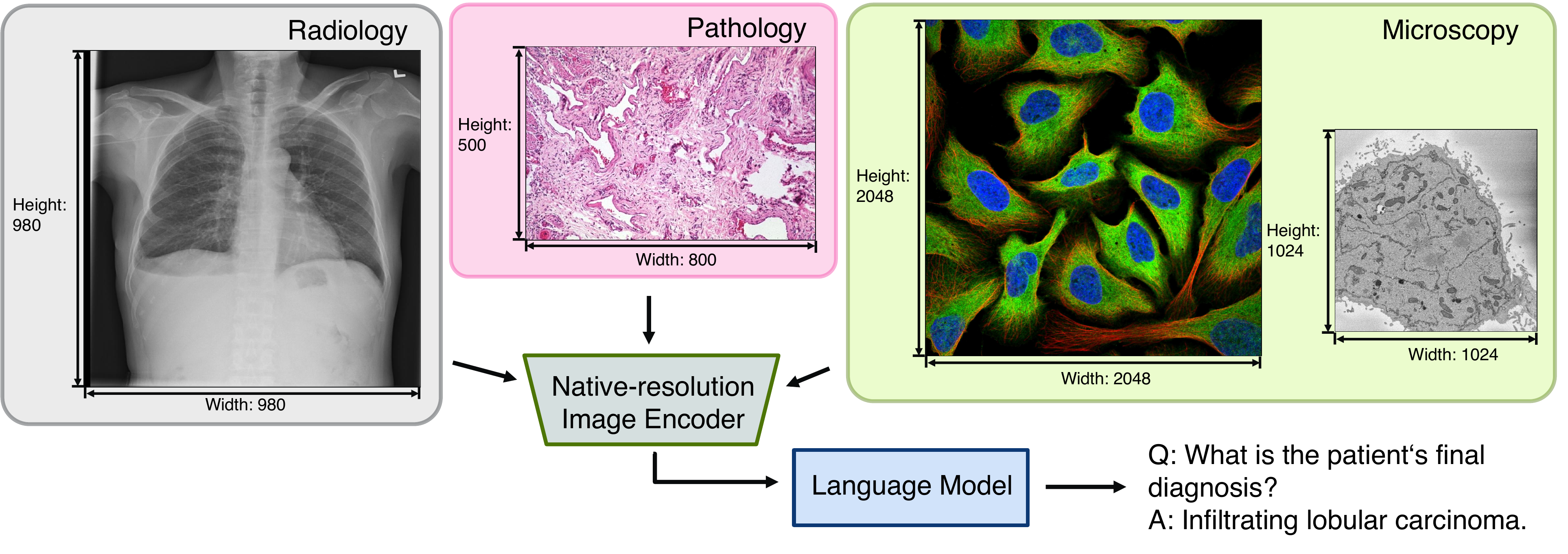}
    \caption{Biomedical MLLMs process images across modalities like radiology, pathology, and microscopy, however standard MLLMs are not designed to handle varying image resolution or high resolutions. This paper argues for native-resolution biomedical MLLMs. The image encoder processes arbitrary-resolution images into tokens dynamically aligned with language tokens, preserving high-frequency details critical for biomedical tasks. The textual output presents an example pathology question-answering task.
    }
  \label{fig:teaser}
\end{figure}

\section{Introduction}
\label{sec:intro}

\begin{figure}[t]
  \centering
  \includegraphics[width=0.4\linewidth]{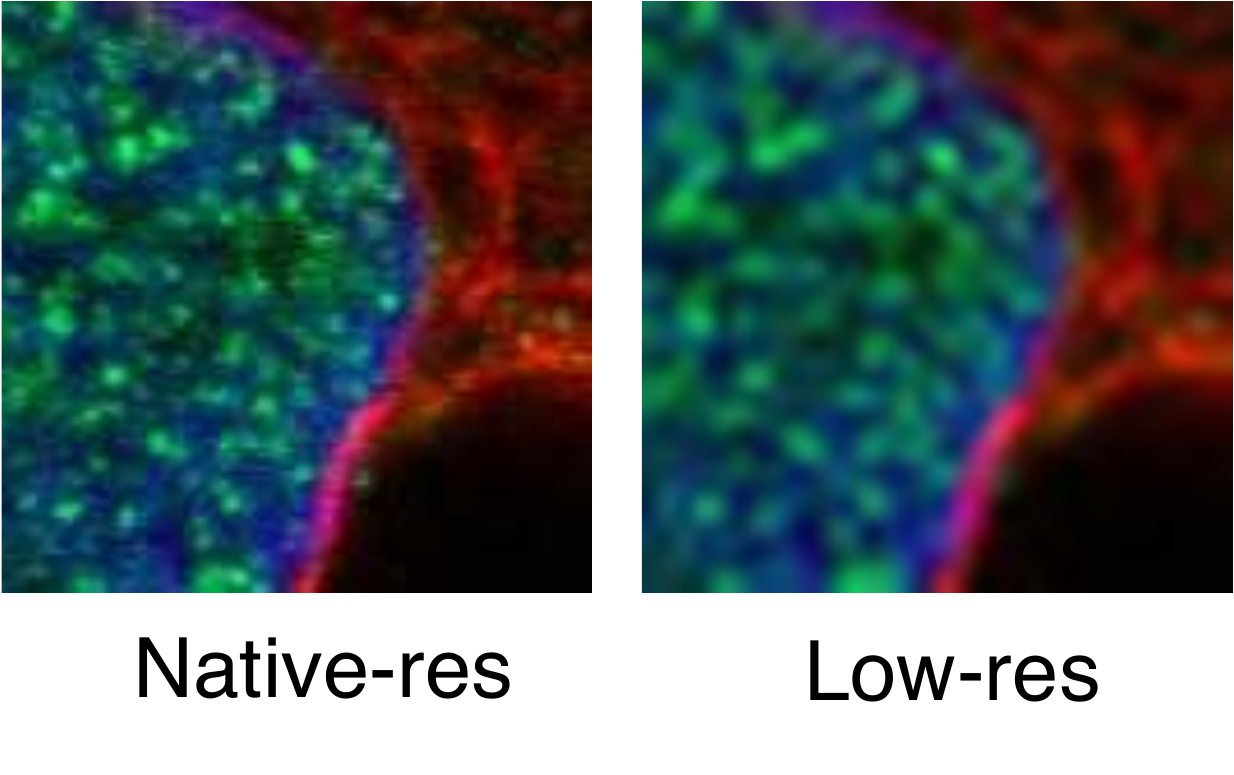}
  \vspace{-5mm}
  \caption{Comparison of image patches at native resolution (left) and low resolution. Downsampling often removes high-frequency details, such as fine textural details in protein staining, that are relevant for accurate interpretation.}
  \label{fig:compare}
\end{figure}

Imaging technologies across a wide spectrum of resolutions are a cornerstone of biomedical research and modern medicine, providing critical insights into biological mechanisms and enabling advanced diagnostics and disease monitoring~\citep{hussain2022modern}. Biomedical imaging spans a wide range of scales and modalities, including chest X-rays~\citep{lau2018dataset}, tissue histopathology~\citep{chen2025wsi}, and fluorescence cell microscopy~\citep{caie2010high, thul2017subcellular}, all of which rely on large images that natively have a high-resolution to capture information at multiple levels. For instance, tumors in CT scans need a broad field of view to determine their anatomical location, alongside native-resolution details to classify tumor subtypes~\citep{dunn2023automated}. Similarly, studying protein function requires a macroscopic view of the entire cell for context and fine-grained resolution to characterize subcellular localization and function~\citep{thul2017subcellular} (\autoref{fig:compare}). Such pathological and microscopic image resolutions range from 2048 by 2048 to more than 12000 by 12000 pixels.

Despite the importance of these native-resolution details, most existing multimodal large language models (MLLMs) are designed to process low-resolution images, such as those commonly found in general-purpose internet datasets~\citep{li2023otter}. Adapting these models to native, high-resolution biomedical applications is challenging. Standard approaches often involve downsampling by approximately an order of magnitude to a fixed low resolution -- this is necessary to enable image patch processing with fixed-size image encoders~\citep{li2024llava, xie2024medtrinity}, but it can obscure critical visual information. Other approaches select patches, omitting the global information~\citep{chen2024dragonfly}.  This raises a key research question: 

\textit{How can image resolution be effectively leveraged during training and inference to preserve the fine-grained features essential for biomedical applications?}

To address this challenge, we investigate the role of resolution fidelity in MLLM performance across tasks where fine-grained visual interpretation is critical. We establish that downsampling biomedical images during training or inference compromises the integrity of the visual data, limiting the utility of MLLMs in biomedical contexts. We explore strategies to mitigate this issue, by proposing to train biomedical MLLMs using architectures that support \textit{native resolution} image encoding \cite{Qwen_VL}, thus naturally modeling image modalities with diverse and high resolutions. To adapt these models to biomedical images, we analyze how resolution impacts model performance and propose  practical solutions for balancing computational complexity with visual details.

More specifically, our experiments demonstrate that when using native-resolution MLLMs during both training and inference significantly improves performance across multiple biomedical tasks, with improvements ranging from 0.54\% to 6.8\% in accuracy across different modalities. We further reveal that misalignment between training and inference resolutions can severely degrade model performance: accuracy drops by up to 48.7\% when using native-resolution training with lower-resolution inference; and accuracy drops up to 43.3\% when using lower-resolution training with native-resolution inference. To address the practical challenges of resolution variability in large-scale biomedical datasets, we propose a mixed-resolution training strategy that effectively maintains performance while accommodating computational constraints, achieving results nearly equivalent to aligned native-resolution training and inference with only a 1.0\% average performance loss. These findings are further validated through zero-shot inference experiments on popular medical VQA benchmarks, where native-resolution inference improves results by 4.0\%.

Based on these findings, we recommend that users of biomedical MLLMs prioritize native-resolution inference when working with models trained with mixed resolutions, and empirically evaluate different inference resolutions when model training details are unknown. For model developers, we advocate implementing balanced mixed-resolution training strategies at the modality level when constructing training datasets, as this approach effectively maintains performance while addressing practical computational constraints. These recommendations aim to optimize the deployment of MLLMs in biomedical applications while preserving the critical fine-grained features necessary for accurate analysis and interpretation.

\subsection*{Generalizable Insights about Machine Learning in the Context of Healthcare}
Our work offers several key insights for machine learning applications in healthcare: (i) We demonstrate the critical importance of image resolution fidelity across multiple biomedical imaging modalities (X-rays, histopathology, microscopy), revealing a consistent pattern where native resolution significantly improves model performance -- challenging the prevalent downsampling paradigm in medical image analysis; (ii) We identify a substantial performance degradation when training and inference resolutions are misaligned, highlighting the need for consistent resolution strategies throughout the ML pipeline; (iii) We propose a practical mixed-resolution training approach that balances computational constraints with performance requirements, achieving results comparable to fully native-resolution methods; and (iv) We provide actionable recommendations for both model users and developers working with high-resolution biomedical images. These findings extend beyond vision applications to establish a broader principle for healthcare ML: preserving the native information density of medical data -- regardless of modality -- is essential for optimal model performance, especially when fine-grained features contain diagnostically relevant information.

\section{Related Work}
\subsection{Image Resolution in Visual Recognition}
Image resolution has been identified as a key attribute of visual recognition. \cite{hao2023understanding}~evaluates the impact of image resolution on object detection. Resolution also affects medical image segmentation performance \citep{rajaraman2023assessing}.  \cite{touvron2019fixing}~raises the train-test resolution discrepancy. They propose an object-resolution-invariant image augmentation technique to counter the effect. Although the augmentation method does not fit modern MLLMs, we identified the resolution discrepancy degrades MLLM performances and proposed solutions.

\subsection{High-resolution MLLMs}
Most existing MLLMs, such as those utilizing pretrained encoders \citep{radford2021learning, liu2023improved, alayrac2022flamingo, li2023otter,li2023blip}, face limitations in their fixed input resolution, often downscaling images to meet computational constraints. This reduction in resolution leads to a loss of critical fine-grained details that are essential for many specialized tasks in cell biology and pathology, such as small object detection, histopathological analysis \citep{gigapath}, and microscopic imaging \citep{lozano2024mu, burgess2025microvqa}.

Several approaches have emerged to address these challenges. Chain-of-Spot \citep{liu2024chain} uses a chain-of-thought prompt templated to localize regions of interest before the user query. In a similar vein, RLogist \citep{zhao2023rlogist} trains a zoom-in strategy by reinforcement learning to localize pathological tissue of interest. While these enhancement methods can be effective, they often struggle to retain global context, add latency, and fail to improve the employed model. End-to-end models, LLaVA-UHD \citep{llava-uhd} and LLaVA-OneVision, \citep{li2024llava} employ a multi-crop strategy to handle high-resolution inputs by splitting large images into smaller segments, thus preserving essential local features while managing computational complexity. Qwen-VL \citep{bai2023qwen}, PaLI-3 \citep{chen2023pali3}, and PaLI-X \citep{chen2023pali} attempt to gradually scale the input resolution of their pretrained encoders, but these approaches often still need to reduce image size, potentially overlooking important visual details at training.  Qwen2-VL \citep{wang2024qwen2} introduces naive dynamic resolution support, allowing the model to flexibly adapt to varying image sizes by employing 2D-RoPE~\citep{heo2024rotary} and post-ViT token compression to limit memory usage and maintain efficiency. We use Qwen2-VL as the base model to adapt to the various resolutions of biomedical images.

\begin{table}[t]
\centering
\label{tab:datasets}
\begin{tabular}{l l l}
\toprule
\textbf{Dataset} & \textbf{Modality} & \textbf{Resolution} \\
\midrule
Subcellular \citep{thul2017subcellular} & Immunofluorescence microscopy & $2048\times2048$  \\
Compound \citep{caie2010high}      & Fluorescence microscopy           & $1280\times1024$  \\
Cervical \citep{hussain2020liquid}  & Liquid-based cytology             & $2048\times1536$  \\
WSI \citep{chen2025wsi}            & Whole-slide pathology             & $\sim3000\times4000$ (0.25$\times$) \\
VQA-RAD \citep{lau2018dataset}     & Radiology                         & $\sim1024\times1024$  \\
PathVQA \citep{he2020pathvqa}       & Pathology                         & $\sim750\times400$  \\
SLAKE \citep{liu2021slake}          & Clinical images (multimodal)      & $\sim1024\times1024$ \\
\bottomrule
\end{tabular}
\caption{Datasets. WSI images are resized because of compute constraints. All other images are in native resolutions.}
\end{table}

\subsection{Biomedical Applications of MLLMs}
Integrating MLLMs in biomedical applications has shown promising advancements, particularly in enhancing the interpretative capabilities across various imaging domains. Models like LLaVA-Med \citep{Li2023LLaVAMedTA} and BiomedGPT \citep{zhang2023biomedgpt} have pioneered efforts to merge medical imaging with scientific textual data, effectively supporting tasks such as diagnosis, visual question answering, and medical report generation. These models build on general-purpose LLMs by introducing specialized biomedical instruction-following datasets \citep{Li2023LLaVAMedTA, xie2024medtrinity}, which enhance their ability to understand domain-specific visual and textual cues.

However, many existing biomedical MLLMs are constrained by limited input resolution. For example, Visual Med-Alpaca \citep{wu2023_visual-med_alpaca} and LLaVA-Med \citep{Li2023LLaVAMedTA} employed CLIP-based image encoders, which are restricted by their native low-resolution capabilities. Such limitations hinder these models' ability to capture detailed biomedical imaging signals -- biomedical MLLM training data has diverse resolutions\cite{lozano2025biomedica} , and tasks often require understanding of microanatomical structures. Recently, Dragonfly \citep{chen2024dragonfly} and Llama3-Med \citep{chen2024advancing} leveraged multi-resolution branches with pretrained vision encoders to support high-resolution biomedical images. However, they still lack flexibility in resolution or fail to scale to very high resolutions (millions of pixels) that are essential to many biomedical applications. Both methods applied hierarchical resolution branches to process the thumbnail image and high-resolution patches, which causes computation overhead by processing redundant visual information. Moreover, none of the prior works studied the impact of inference resolution on performance.

\section{Experiments}

Our experiments focus on classification and Visual Question Answering (VQA) tasks, where an MLLM is provided with an image and a text-based question and must generate an accurate answer. We evaluate performance across seven tasks from three representative biomedical imaging modalities (radiology, pathology, microscopy), chosen for their reliance on native-resolution data and their importance in both research and clinical practice. For clarity in the figures and tables, we denote these datasets as ``Subcellular'', ``Compound'', ``Cervical'', ``WSI'', ``VQA-RAD'', ``PathVQA'', ``SLAKE'' (\autoref{tab:datasets}). Leveraging the QwenVL-2 architecture \citep{wang2024qwen2}, which dynamically processes images of varying resolutions by splitting them into fixed-size patches (\autoref{fig:teaser}), we preserve rich visual details with reasonable computational resources. 

Through this study, we aim to highlight the critical importance of native-resolution image processing in the design and application of biomedical MLLMs, while offering practical recommendations to optimize their performance.
Our findings highlight the critical role of native-resolution images in advancing biomedical MLLMs. First, we demonstrate that native-resolution training significantly improves performance across multiple biomedical classification and visual question-answering tasks. Second, we establish that alignment between training and inference resolutions is crucial, as misalignment leads to substantial performance degradation. To address the practical challenge posed by resolution variability in large-scale biomedical datasets, we propose mixed-resolution training, which effectively mitigates misalignment issues while preserving the benefits of native-resolution inference. Based on these insights, we recommend that future biomedical MLLMs prioritize native-resolution inference and that training datasets incorporate a balanced mix of resolutions to maximize performance and generalizability. These principles are essential for optimizing MLLMs to meet the demands of fine-grained biomedical image analysis.

\subsection{Data}

\begin{figure}[t]
  \centering
  \includegraphics[width=0.6\linewidth]{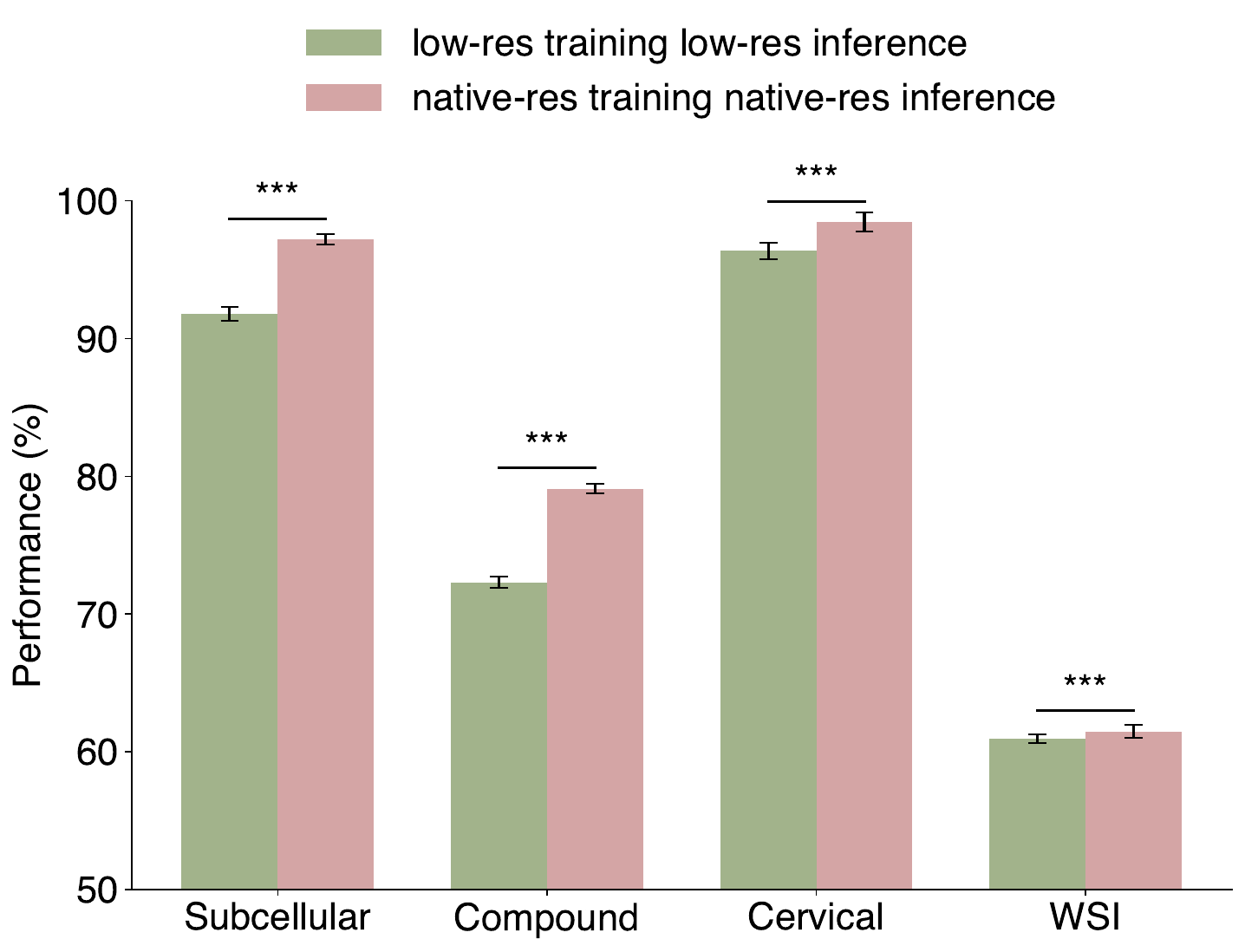}
  \caption{Performance comparison of native-resolution and low-resolution training and inference. Native-resolution training and inference achieve superior results, emphasizing the importance of resolution fidelity.}
  \label{fig:low_high_bar_plot_limited}
\end{figure}

To comprehensively evaluate the performance of multimodal large language models (MLLMs) in biomedical applications, we conducted experiments on seven diverse datasets that span critical imaging modalities and biomedical tasks (\autoref{tab:datasets}). 

\textbf{- Subcellular, \citet{thul2017subcellular}
}
This dataset contains immunofluorescence microscopy images of human cells, annotated for subcellular protein localization~. It provides native, high-resolution images essential for studying protein function and cellular context. We measure the cell line classification accuracy on this dataset.

\textbf{- Compound, \citet{caie2010high}
}
This dataset focuses on fluorescence microscopy-based high-content screening for compound profiling. Images capture cellular responses to chemical perturbations, requiring fine-grained details for accurate classification.  We measure the compound profiling classification accuracy on this dataset.

\textbf{- Cervical, \citet{hussain2020liquid}
}
This dataset includes native-resolution images from liquid-based cytology, annotated for pre-cancerous and cervical cancer lesions. The dataset is critical for evaluating diagnostic capabilities in cervical cytology.  We measure the lesion classification accuracy on this dataset.

\textbf{- WSI, \citet{chen2025wsi}
}
This dataset comprises whole-slide pathology images annotated for tasks such as tumor classification and diagnostic visual question answering. These large, native-resolution images capture essential morphological details across broad tissue areas. Because the original resolution is too high to fit in the memory of a single server, we downsample it to be around 25\% of the original resolution, which is a bout 3000*4000. We measure the accuracy of the closed-form question-answering on this dataset.

\textbf{- VQA-RAD, \citet{lau2018dataset}
}
VQA-RAD is a medical VQA benchmark dataset containing radiology images paired with clinically relevant questions. The questions cover diverse topics, such as anatomical structures and disease diagnosis, requiring both visual and textual understanding. We measure the closed-set question-answering accuracy on this dataset.

\textbf{- PathVQA, \citet{he2020pathvqa}
}
PathVQA is a VQA dataset based on pathology images, designed to test a model’s ability to answer questions about cellular and tissue-level features. The dataset emphasizes the need for native-resolution image processing. We measure the closed-set question-answering accuracy on this dataset.

\textbf{- SLAKE, \citet{liu2021slake}
}
SLAKE is another medical VQA dataset that focuses on multimodal reasoning over clinical images and associated textual data. Its questions demand fine-grained visual interpretation alongside contextual understanding. We measure the closed-set question-answering accuracy on the English subset of this dataset.

\begin{table}[t]
    \centering
    \footnotesize 
    \caption{Resolution alignment in training and inference for Qwen2-VL 2B. Mixed-resolution training enables models to adapt effectively to both low-, native-, and mixed-resolution inference.}
    \label{tab:performance}
    \begin{tabular}{llccccccc}
        \toprule
        \textbf{Training} & \textbf{Inference} & \textbf{Subcellular} & \textbf{Compound} & \textbf{Cervical} & \textbf{WSI} & \textbf{VQA-RAD} & \textbf{PathVQA} & \textbf{SLAKE} \\
        \midrule
        \multirow{3}{*}{No} 
            & Low-res     & 1.80 & 8.60 & 21.35 & 14.55 & 39.25 & 16.96 & 49.09 \\
            & Native-res  & 3.85 & 9.10 & 22.63 & 18.36 & 39.65 & 18.19 & 48.71 \\
            & Mixed-res   & 2.86 & 8.84 & 21.94 & 16.45 & 39.43 & 17.52 & 48.92 \\
        \midrule
        \multirow{3}{*}{Low-res } 
            & Low-res     & 91.80 & 72.30 & 96.37 & 60.95 & 30.93 & 20.81 & 56.63 \\
            & Native-res  & 64.35 & 29.00 & 82.90 & 58.91 & 29.58 & 19.20 & 52.48 \\
            & Mixed-res   & 77.68 & 50.45 & 89.46 & 59.85 & 30.32 & 20.04 & 54.57 \\
        \midrule
        \multirow{3}{*}{Native-res } 
            & Low-res     & 48.50 & 46.30 & 80.83 & 54.45 & 38.54 & 18.65 & 54.74 \\
            & Native-res  & 97.20 & 79.10 & 98.45 & 61.49 & 42.49 & 24.88 & 60.23 \\
            & Mixed-res   & 72.85 & 62.70 & 89.64 & 58.06 & 40.56 & 21.75 & 57.48 \\
        \midrule
        \multirow{3}{*}{Mixed-res } 
            & Low-res     & 78.55 & 58.80 & 89.24 & 56.93 & 39.35 & 20.45 & 55.98 \\
            & Native-res  & 95.20 & 78.00 & 97.93 & 61.34 & 41.92 & 22.94 & 58.03 \\
            & Mixed-res   & 90.48 & 72.65 & 95.31 & 60.24 & 41.08 & 22.15 & 57.36 \\
        \bottomrule
    \end{tabular}
\end{table}

\begin{table}[t]
    \centering
    \footnotesize 
    \caption{Resolution alignment in training and inference for InternVL2.5 2B. Mixed-resolution training enables models to adapt effectively to both low-, native-, and mixed-resolution inference.}
    \label{tab:internvl_performance}
    \begin{tabular}{llccccccc}
        \toprule
        \textbf{Training} & \textbf{Inference} & \textbf{Subcellular} & \textbf{Compound} & \textbf{Cervical} & \textbf{WSI} & \textbf{VQA-RAD} & \textbf{PathVQA} & \textbf{SLAKE} \\
        \midrule
        \multirow{3}{*}{No} 
            & Low-res     & 2.15 & 9.45 & 23.86 & 16.12 & 41.75 & 18.45 & 52.18 \\
            & Native-res  & 4.32 & 10.28 & 25.19 & 20.43 & 42.06 & 19.84 & 51.92 \\
            & Mixed-res   & 3.29 & 9.82 & 24.53 & 18.25 & 41.89 & 19.14 & 52.05 \\
        \midrule
        \multirow{3}{*}{Low-res } 
            & Low-res     & 94.38 & 75.64 & 97.25 & 63.47 & 32.85 & 22.54 & 59.21 \\
            & Native-res  & 68.74 & 31.56 & 84.68 & 61.35 & 31.42 & 20.86 & 55.13 \\
            & Mixed-res   & 81.53 & 53.20 & 90.92 & 62.41 & 32.15 & 21.68 & 57.32 \\
        \midrule
        \multirow{3}{*}{Native-res } 
            & Low-res     & 52.64 & 49.85 & 83.47 & 57.28 & 40.96 & 20.34 & 57.42 \\
            & Native-res  & 98.45 & 82.36 & 99.12 & 64.23 & 45.18 & 26.73 & 63.15 \\
            & Mixed-res   & 75.32 & 66.15 & 91.26 & 60.74 & 42.95 & 23.47 & 60.18 \\
        \midrule
        \multirow{3}{*}{Mixed-res } 
            & Low-res     & 82.16 & 62.47 & 92.38 & 59.84 & 41.87 & 22.18 & 58.76 \\
            & Native-res  & 97.35 & 81.24 & 98.64 & 64.05 & 44.63 & 24.85 & 60.94 \\
            & Mixed-res   & 93.67 & 76.83 & 97.02 & 63.28 & 43.95 & 24.08 & 60.28 \\
        \bottomrule
    \end{tabular}
\end{table}

\begin{table}[t]
    \centering
    \footnotesize 
    \caption{Resolution alignment in training and inference for LLaVA-OneVision 2B. Mixed-resolution training enables models to adapt effectively to both low-, native-, and mixed-resolution inference.}
    \label{tab:llava_performance}
    \begin{tabular}{llccccccc}
        \toprule
        \textbf{Training} & \textbf{Inference} & \textbf{Subcellular} & \textbf{Compound} & \textbf{Cervical} & \textbf{WSI} & \textbf{VQA-RAD} & \textbf{PathVQA} & \textbf{SLAKE} \\
        \midrule
        \multirow{3}{*}{No} 
            & Low-res     & 1.45 & 7.23 & 18.76 & 12.38 & 34.85 & 14.27 & 43.65 \\
            & Native-res  & 3.12 & 7.84 & 19.85 & 15.67 & 35.42 & 15.63 & 43.18 \\
            & Mixed-res   & 2.34 & 7.58 & 19.30 & 14.06 & 35.15 & 14.94 & 43.47 \\
        \midrule
        \multirow{3}{*}{Low-res } 
            & Low-res     & 82.75 & 65.48 & 89.75 & 54.32 & 27.84 & 18.26 & 50.15 \\
            & Native-res  & 58.63 & 24.75 & 76.53 & 52.45 & 26.73 & 16.85 & 47.29 \\
            & Mixed-res   & 70.42 & 45.13 & 83.16 & 53.40 & 27.31 & 17.63 & 48.74 \\
        \midrule
        \multirow{3}{*}{Native-res } 
            & Low-res     & 43.27 & 40.65 & 74.38 & 48.93 & 34.62 & 16.48 & 49.35 \\
            & Native-res  & 88.64 & 71.85 & 91.26 & 55.74 & 38.75 & 21.95 & 54.87 \\
            & Mixed-res   & 65.85 & 56.24 & 82.91 & 52.35 & 36.58 & 19.16 & 52.13 \\
        \midrule
        \multirow{3}{*}{Mixed-res } 
            & Low-res     & 70.86 & 51.74 & 82.65 & 50.87 & 35.48 & 18.17 & 50.76 \\
            & Native-res  & 87.35 & 70.94 & 90.84 & 55.28 & 37.94 & 20.65 & 53.25 \\
            & Mixed-res   & 83.64 & 65.32 & 88.75 & 54.16 & 37.31 & 19.94 & 52.48 \\
        \bottomrule
    \end{tabular}
\end{table}

\begin{table}[t]
    \centering
    \footnotesize 
    \caption{Resolution alignment in training and inference for Qwen2-VL 7B. Mixed-resolution training enables models to adapt effectively to both low-, native-, and mixed-resolution inference.}
    \label{tab:qwen2vl_7b_performance}
    \begin{tabular}{llccccccc}
        \toprule
        \textbf{Training} & \textbf{Inference} & \textbf{Subcellular} & \textbf{Compound} & \textbf{Cervical} & \textbf{WSI} & \textbf{VQA-RAD} & \textbf{PathVQA} & \textbf{SLAKE} \\
        \midrule
        \multirow{3}{*}{No} 
            & Low-res     & 2.46 & 10.24 & 23.59 & 16.82 & 42.57 & 19.34 & 52.64 \\
            & Native-res  & 4.93 & 11.35 & 25.41 & 21.52 & 43.24 & 20.85 & 52.18 \\
            & Mixed-res   & 3.78 & 10.93 & 24.53 & 19.37 & 42.95 & 20.12 & 52.43 \\
        \midrule
        \multirow{3}{*}{Low-res } 
            & Low-res     & 96.37 & 78.54 & 98.25 & 65.42 & 33.85 & 32.36 & 61.27 \\
            & Native-res  & 69.28 & 33.76 & 85.62 & 63.45 & 32.38 & 30.82 & 56.84 \\
            & Mixed-res   & 82.48 & 56.32 & 92.14 & 64.37 & 33.13 & 31.65 & 59.05 \\
        \midrule
        \multirow{3}{*}{Native-res } 
            & Low-res     & 53.42 & 52.37 & 84.76 & 59.78 & 41.95 & 35.82 & 58.93 \\
            & Native-res  & 98.45 & 85.23 & 99.35 & 67.26 & 46.78 & 40.47 & 64.85 \\
            & Mixed-res   & 76.32 & 68.94 & 92.18 & 63.52 & 44.32 & 38.24 & 61.92 \\
        \midrule
        \multirow{3}{*}{Mixed-res } 
            & Low-res     & 83.62 & 64.85 & 92.75 & 61.34 & 42.85 & 33.75 & 60.38 \\
            & Native-res  & 97.83 & 83.45 & 98.83 & 66.85 & 45.95 & 38.67 & 62.73 \\
            & Mixed-res   & 93.47 & 78.38 & 97.25 & 65.24 & 44.86 & 37.23 & 61.85 \\
        \bottomrule
    \end{tabular}
\end{table}

\subsection{Experiment Settings}
We use Qwen2-VL 2B \citep{wang2024qwen2} as a base model unless otherwise specified, with no architectural change.
Qwen2-VL was chosen for two reasons: (1) it handles arbitrary image resolution by mapping image patches into a dynamic number of image tokens that merge with language tokens, balancing performance and compute efficiency; (2) it was the best-performing MLLM in general domains at the time of experiments. The model has a simple architecture orchestrating a native-resolution image encoder and a language model, as illustrated in Figure 2 of \cite{wang2024qwen2}.

In a common vision-language modeling approach, the Qwen2-VL model employs a language-pretrained large language model for further vision-language pretraining and instruction tuning. The vision-language training data details are not revealed.

Our finetuning on biomedical data was done with controlled hyperparameters. The total batch size is 128. The micro-batch size and gradient accumulation steps vary to control the total batch size. The learning rate is 1e-5 with a cosine scheduler. The weight decay is 0.1. The experiments are performed with NVIDIA A100 and A6000 graphic processing units.

All the models are fully finetuned with bf16 and DeepSpeed Zero2. The training time ranged from 10 to 120 A100 GPU hours, depending on the resolution and image quantities. We compared full finetuning (38.54\%, training for 8 A100 hours) with LoRA (30.22\%, training for 4.8 A100 hours) on VQA-RAD when we started the experiments, native-res training low-res inference. We choose full finetuning for the rest of the experiments because it performs better with manageable computing expenditure. 

\subsection{Impact of Native Resolution on Model Performance}

\begin{figure}[t]
  \centering
  \includegraphics[width=0.5\linewidth]{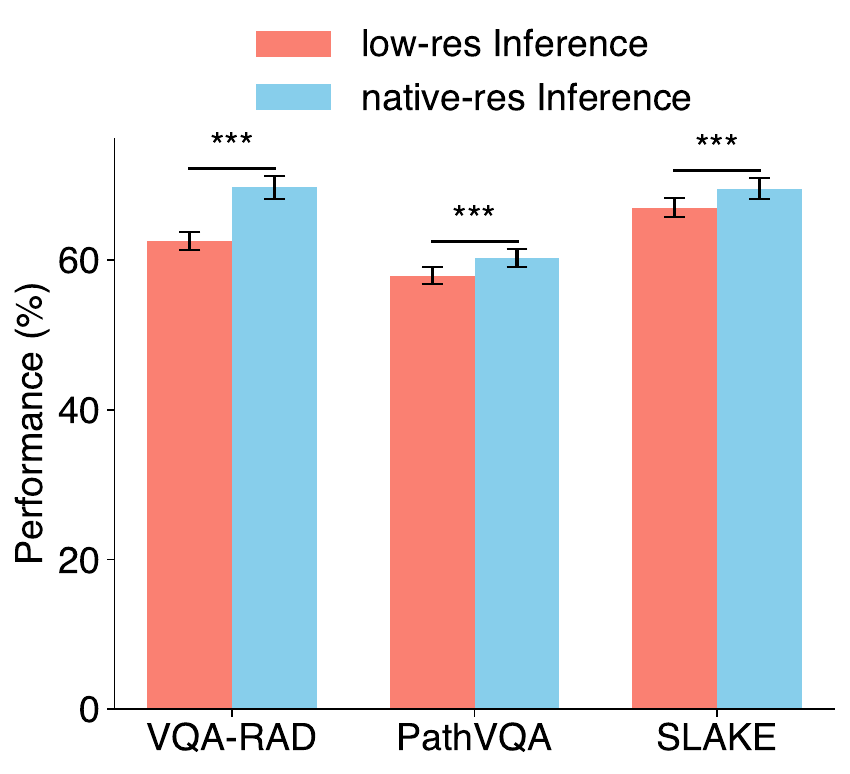}
  \caption{Native-resolution inference improves off-the-shelf models. Qwen2-VL 7B shows consistent gains on radiological and pathological question-answering tasks with native-resolution input.}
  \label{fig:inference_performance_comparison}
\end{figure}

Our first finding establishes that performing training and inference with native-resolution images improves performance on biomedical VQA tasks. We fine-tune the base MLLM in native or low resolution, and performs inference with the same resolution. The results are in \autoref{fig:low_high_bar_plot_limited}. Native-resolution training improves immunofluorescence microscopy cell~\citep{thul2017subcellular} classification accuracy by 5.4\%, $P<0.001$; fluorescence microscopy compound-profiling~\citep{caie2010high} classification accuracy by 6.8\%, $P<0.001$; pre-cancerous and cervical cancer lesion~\citep{hussain2020liquid} classification accuracy by 2.08\%, $P<0.001$.; whole-slide pathology~\citep{chen2025wsi} visual question answering accuracy by 0.54\%, $P<0.001$. These results support the intuition that image processing tasks requiring fine details need native-resolution images, suggesting that biomedical MLLMs should incorporate native-resolution images.

\subsection{Resolution Alignment Between Training and Inference}

Our second finding reveals that misalignment between training and inference resolutions can substantially degrade performance in biomedical classification/VQA tasks. When we use a model trained with native resolution but perform inference with lower resolution, we observe severe performance degradation: -48.7\%, -32.8\%, -17.6\% and -7.0\% across tasks (\autoref{tab:performance}). Similarly, training with lower resolution and testing on native resolution leads to significant performance losses of -27.4\%, -43.3\%, -13.5\% and -2.0\%. Notably, these misaligned configurations perform worse than consistently using lower resolution for both training and inference, indicating that resolution alignment between training and testing is more crucial than the actual resolution used for inference. Decreasing inference resolution leads to more severe performance than the other way around. We attribute it to the task difficulty gap between training and inference. Training with native resolution and inference with low resolution imposes a more difficult task that the model didn't learn at training.

However, these findings present a practical challenge: biomedical MLLMs typically train on large datasets from various sources with different image resolutions, making strict training-inference resolution alignment impractical. To address this challenge, we investigated mixed-resolution training, where 50\% of training samples use lower resolution. Our results in \autoref{tab:performance} show that mixed-resolution training effectively mitigates the problems of misaligned train-test resolutions. We observe consistent results on other models like InternVL2.5 2B (\autoref{tab:internvl_performance}),  LLaVA-OneVision 2B (\autoref{tab:llava_performance}), and larger models like Qwen2-VL 7B (\autoref{tab:qwen2vl_7b_performance}). When using high-resolution inference with mixed-resolution training, performance nearly matches that of aligned native-resolution training and inference, with only a 1.0\% average performance loss. Furthermore, with mixed-resolution training, native-resolution inference consistently outperforms lower-resolution inference by an average margin of 12.2\%. 
Notably, the compute cost of MLLM increases proportionally with the number of pixels. Thus mixed-resolution training also balances between compute efficiency and performance requirements. 
For model developers, we recommend implementing balanced mixed-resolution training strategies, as our results show that while training-inference misalignment can be catastrophic, 50-50 mixed-resolution training effectively maintains performance. We suggest implementing this balance at the level of each imaging modality when constructing training datasets.

\subsection{Inference Strategy}

Based on the previous findings, we offer two practical recommendations. For users inferencing with biomedical MLLMs, we recommend using native-resolution inference when working with models known to be trained with mixed resolutions. If the model's training resolution details are unknown, users should empirically evaluate different inference resolutions to determine the optimal setting. We validate this approach through zero-shot inference experiments with a pretrained Qwen2-VL 7B model on standard medical VQA benchmarks: VQA-RAD~\citep{lau2018dataset}, PathVQA~\citep{he2020pathvqa}, and SLAKE~\citep{liu2021slake}. These experiments confirm that inference resolution significantly affects performance when the training resolution of a massive pre-trained model is unknown, with native resolution improving results by 4.0\% on average (\autoref{fig:inference_performance_comparison}).

\begin{figure*}[t]
\floatconts
  {fig:resizing_bar_plot}
  {\caption{Comparing low-resolution, high-resolution, and images resampled from low-resolution images.}}
  {\includegraphics[width=\linewidth]{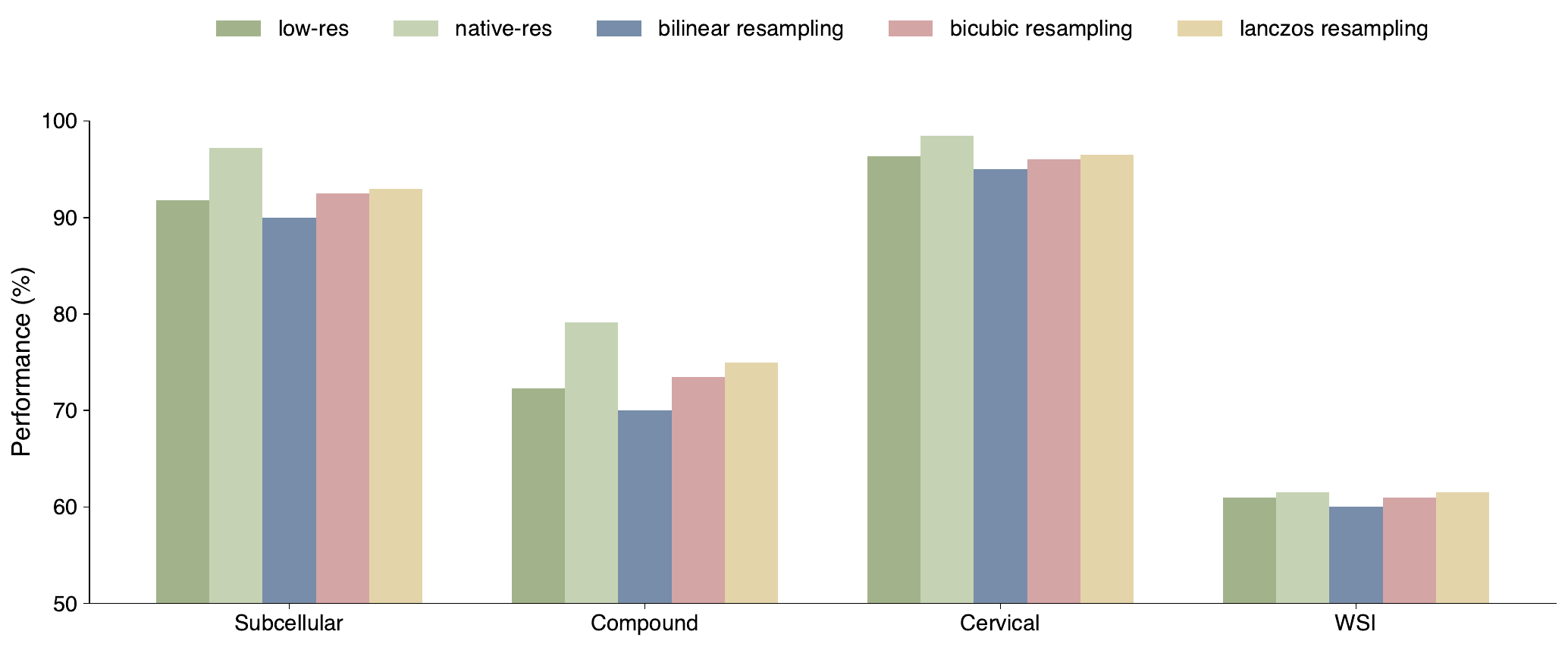}}
\end{figure*}

\subsection{Resizing}

In order to ablate the effect of image resolution, we compare images upsampled to the native resolution from low-resolution images, as shown in \autoref{fig:resizing_bar_plot}.
Our analysis of different resampling techniques reveals that advanced resampling methods can partially mitigate the performance degradation caused by resolution reduction, though they cannot fully restore native-resolution performance. Lanczos resampling consistently outperforms other methods across all tasks, achieving 93.1\% accuracy on subcellular classification compared to the native-resolution performance of 97.2\%, and 75.2\% on compound classification versus 79.1\% with native resolution. Bicubic resampling shows moderate improvements over bilinear resampling, with performance gains of 2.5\% and 3.3\% on subcellular and compound classification tasks, respectively. Notably, the effectiveness of resampling methods varies across different imaging modalities, with the greatest benefits observed in tasks requiring fine-grained feature preservation, such as subcellular and compound analysis, while showing minimal impact on whole-slide imaging tasks where the performance gap between resampled and native resolution remains relatively constant.

Based on our experimental results, we recommend a hierarchical approach to resolution handling in biomedical MLLMs. When data access permits, native resolution should be maintained as it consistently delivers superior performance across all tasks. In resource-constrained environments where a user only has access to low-resolution images, Lanczos resampling should be prioritized as the preferred downsampling method, as it recovers 95.7\% of native-resolution performance on average across all tasks. 

\section{Discussion} 
In this study, we present two key findings about image resolution in biomedical MLLMs, followed by practical recommendations for implementation. First, we demonstrate that native-resolution images improve performance across multiple biomedical visual question-answering (VQA) tasks. Second, we show the critical importance of alignment between training and inference resolutions. These findings lead us to important practical considerations for both users and developers of biomedical MLLMs.

\paragraph{Broader Impact} The findings from this study have significant implications for the deployment of MLLMs in healthcare settings, potentially improving diagnostic accuracy and research outcomes across various biomedical imaging modalities. By establishing best practices for resolution handling in biomedical MLLMs, our work could accelerate the development of more reliable artificial intelligence systems for medical image analysis, potentially leading to earlier disease detection and more accurate diagnoses. However, the computational resources required for native-resolution processing may limit accessibility to well-resourced institutions, potentially exacerbating healthcare disparities. Additionally, improved performance of these systems could lead to over-reliance on automated analysis, emphasizing the importance of maintaining human oversight and using these tools as aids rather than replacements for clinical expertise.

\paragraph{Limitations and Future Work} While our study demonstrates the importance of resolution fidelity in biomedical MLLMs, several limitations should be acknowledged. First, our experiments primarily focus on specific imaging modalities and may not generalize to all types of biomedical imaging. The computational demands of native-resolution processing also present practical constraints for real-time applications and resource-limited settings. Furthermore, our mixed-resolution training strategy, while effective, may not be optimal for all scenarios, and the ideal ratio of resolution mixing might vary across different applications and imaging modalities. Future work should explore more efficient architectures for handling multi-resolution inputs and investigate adaptive resolution selection mechanisms based on task-specific requirements and computational constraints.

\paragraph{Acknowledgement} We thank Xiaohan Wang, Alejandro Lozano, Anita Rau, and Sanket Gupte for their thoughtful discussions. 

\newpage

\bibliography{sample}
\newpage
\appendix
\section{Data and Code Availability}
All the datasets are available online from the original sources:

\textbf{- Subcellular, \citet{thul2017subcellular}
}
Immunofluorescence microscopy images of human cells, annotated for subcellular protein localization.

\textbf{- Compound, \citet{caie2010high}
}
Fluorescence microscopy-based high-content screening for compound profiling.

\textbf{- Cervical, \citet{hussain2020liquid}
}
Native-resolution images from liquid-based cytology, annotated for pre-cancerous and cervical cancer lesions.

\textbf{- WSI, \citet{chen2025wsi}
}
Whole-slide pathology images annotated for tasks such as tumor classification and diagnostic visual question answering. 

\textbf{- VQA-RAD, \citet{lau2018dataset}
}
A medical VQA benchmark dataset containing radiology images paired with clinically relevant questions.

\textbf{- PathVQA, \citet{he2020pathvqa}
}
A VQA dataset based on pathology images, designed to test a model’s ability to answer questions about cellular and tissue-level features.

\textbf{- SLAKE, \citet{liu2021slake}
}
A medical VQA dataset, focusing on multimodal reasoning over clinical images and associated textual data.

The third-party code for training is available on GitHub (\url{https://github.com/modelscope/ms-swift}). 
The evaluation code is available on (\url{https://github.com/cliangyu/med_eval}).

\paragraph*{Institutional Review Board (IRB)}
This study uses public datasets under the data license. The original authors previously de-identified any patient-derived images in compliance with applicable privacy laws and institutional guidelines. The Institutional Review Board \href{https://researchcompliance.stanford.edu/panels/hs/for-all-researchers}{guidelines} were reviewed, and the public use of deidentified images does not constitute human subjects research.

\end{document}